# Block DCT filtering using vector processing

Mostafa Amin-Naji
Faculty of Electrical and Computer Engineering
Babol Noshirvani University of Technology
Babol, Iran
Mostafa.Amin.Naji@gmail.com

Ali Aghagolzadeh
Faculty of Electrical and Computer Engineering
Babol Noshirvani University of Technology
Babol, Iran
Aghagol@nit.ac.ir

*Abstract*—Filtering is an important issue in signals and images processing. Many images and videos are compressed using discrete cosine transform (DCT). For reducing the computation complexity, we are interested in filtering block and images directly in DCT domain. This article proposed an efficient and yet very simple filtering method directly in DCT domain for any symmetric, asymmetric, separable, inseparable and one or two dimensional filter. The proposed method is achieved by mathematical relations using vector processing for image filtering which it is equivalent to the spatial domain zero padding filtering. Also to avoid the zero padding artifacts around the edge of the block, we prepare preliminary matrices in DCT domain by implementation elements of selected mask which satisfies border replication for a block in the spatial domain. To evaluate the performance of the proposed algorithm, we compared the spatial domain filtering results with the results of the proposed method in DCT domain. The experiments show that the results of our proposed method in DCT are exactly the same as the spatial domain filtering.

*Keywords—DCT; block filtering; vector processing; image filtering*

I. INTRODUCTION

In image/video processing, filtering and manipulation images and videos are common and have widely usage. The discrete cosine transform is used mainly in image and video compression such as JPEG [1], MPEG [2] and H.261 [3]. In order to filter these images and videos, the compressed images and video should be transformed into spatial domain. After filtering process, the results must be transformed back into the compressed domain. Filtering data in the compressed (DCT) domain reduces the computation cost. Also the block DCT filtering is very common and popular in image processing approaches like noise removal, sharpening and de-blurring. For example in some multi-focus image fusion methods, we need to convolve a mask of low pass filter on the block directly in DCT domain [4, 5]. Thus the main goal of this paper is image filtering directly in DCT (compressed) domain. For discrete Fourier transform (DFT) there is convolution-multiplication property (CMP). In DFT, the linear 2-D convolution of two matrices is the same as coefficient to coefficient multiplication of the corresponding elements; whereas there isn't such clear and simple manipulation property in DCT domain.

Numerous different studies have been conducted on DCT filtering. In the early researches, Chen and Fralick [6] revealed the coefficient to coefficient multiplication in DCT domain can derive by circular convolution. Ngan and Clarke [7] used this property for low pass filtering and Chitprasert and Rao [8] showed an effective simplification of CMP which proposed in [7]. But these methods worked only for circular convolution which introduces undesirable artifacts of block edge. Martucci [9] introduced symmetric convolution for discrete trigonometric transform (DTT) and related it to DCT and discrete sine transform (DST). The linear convolution can be performed in this method instead of DFT and circular convolution if the input block is zero padded before transformation. So this method needs zero padding and also has complex preprocessing for the selected filter. Kresch and Merhav introduced DCT domain filtering method for 8×8 blocks [10]. This method needs DST coefficients along with DCT coefficients of the corresponding block. When the DST coefficients are unavailable, it must be computed from the existing DCT coefficients which requires more computing. Also because of transformation from cosine to sine and sine to cosine, this method has complex signal flow.

Yim in [11] introduced a method for separable symmetric linear filtering in DCT domain which needs less number of multiplications. But this method is appropriate when the selected filter coefficients are symmetric and it cannot be implemented for asymmetric filters. In [12] Reju et al. introduced a circular convolution property in DST and DCT. Linear convolution can implement here by zero padding the input data. This method is not dependent on whether the filter is symmetric or asymmetric, but it works only for one dimensional convolution. Suresh and Sreenivas in [13] implemented linear filtering for DCT/DST using convolution-multiplication property of DTT and extended it for modified DCT and modified DST (MDCT/MDST). This method designed for one dimensional filter. Also, filter implementation is near to exact and approximate for symmetric and asymmetric filters, respectively. Viswanath et al. in [14] used symmetric convolution for separable linear block filtering in DCT domain. This method uses some matrices derived by satisfying some symmetric and sparse properties which reduces the computation costs. None of the above mentioned methods is suitable when the selected filter is two dimensional, inseparable or asymmetric. Also for each selected filter, these methods require complex calculation for preparing the primary filtering computational instruments.

In this paper a simple but comprehensive approach for block filtering, directly in DCT domain, is presented. The proposed method has no limitation for one or two dimensions, symmetric or asymmetric and separable or inseparable filters. For any size of the desired filter, we can simply implement the mask in the form of matrix with the same size as the block to be filtered. In order to filter the block, the obtained matrices are used to multiply directly in DCT domain by vector processing. The result of the convolved block is exactly the same as the spatial domain result in Uint8 format. The primary obtained matrices for filtering process can be easily reconstructed for any small change of mask filter without complex calculation. Also, our method is composed of simple and fixed operations (multiplication and summation of the same size matrices) which would be appropriate for efficient VLSI implementation.

## II. POPOSED METHOD

### A. Discerete Cosin Transform In Vector Procrssing

Two-dimensional DCT transform of N×N block (m, n) of the image is defined using vector processing as (1):

$$BLOCK = C. block. C^t \quad (1)$$

where C is an orthogonal matrix consisting of the cosine coefficients and $C^t$ the transpose of C. Also, we have:

$$C^{-1} = C^t \quad (2)$$

BLOCK is the DCT coefficient representation of block. The inverse DCT of BLOCK is defined as (3):

$$block = C^t. BLOCK. C \quad (3)$$

### B. Filtering In DCT Domain

For a filtering process, a custom mask is intended to be performed on the block. This filter (mask) has no restriction such as symmetrical, separable and two-dimension. Due to limitation in article pages, the proposed method is explained with implementation 3×3 custom mask on a 8×8 block. Also the 8×8 blocks are very common in image processing approaches like JPEG. The proposed method can be extended to any larger size of the mask of filter and block. We define a custom mask filter W as (4):

| w11 | w12 | w13 |
|-----|-----|-----|
| w21 | w22 | w23 |
| w31 | w32 | w33 |

W=  (4)

Filtered block with mask W is achieved by (5) in the spatial domain:

$$f(i, j) = \sum_{k=-1}^{+1} \sum_{l=-1}^{+1} w(l, k) \times block(i - k, j - l) \quad (5)$$

In order to convolve the mask on the block directly in DCT domain, we simulate convolution operation with multiplication the matrices (the same size of block) for every row or column of the mask. The results must satisfy the zero padding filtering in the spatial domain. The developed matrices are transferred to DCT domain just one time after designing for every selected mask and then are used in block filtering process as constant matrices. The implementation procedure is divided into three parts: multiplication of the first, the second and the third row (or column) of mask on the block. For implementation the first row of mask on the block, w11, w12 and w13 should slide on the block; so we define matrix "b" that must be multiplied on the block. Since the first row of mask is not related to the first row of block, the first row of block×b should be zero. This operation is performed by developing matrix "a" that is multiplied on the block (block×b).

$$a = \begin{pmatrix} 0 & 0 & 0 & 0 & 0 & 0 & 0 & 0 \\ 1 & 0 & 0 & 0 & 0 & 0 & 0 & 0 \\ 0 & 1 & 0 & 0 & 0 & 0 & 0 & 0 \\ 0 & 0 & 1 & 0 & 0 & 0 & 0 & 0 \\ 0 & 0 & 0 & 1 & 0 & 0 & 0 & 0 \\ 0 & 0 & 0 & 0 & 1 & 0 & 0 & 0 \\ 0 & 0 & 0 & 0 & 0 & 1 & 0 & 0 \\ 0 & 0 & 0 & 0 & 0 & 0 & 1 & 0 \end{pmatrix}_{8\times 8}$$

$$b = \begin{pmatrix} w12 & w11 & 0 & 0 & 0 & 0 & 0 & 0 \\ w13 & w12 & w11 & 0 & 0 & 0 & 0 & 0 \\ 0 & w13 & w12 & w11 & 0 & 0 & 0 & 0 \\ 0 & 0 & w13 & w12 & w11 & 0 & 0 & 0 \\ 0 & 0 & 0 & w13 & w12 & w11 & 0 & 0 \\ 0 & 0 & 0 & 0 & w13 & w12 & w11 & 0 \\ 0 & 0 & 0 & 0 & 0 & w13 & w12 & w11 \\ 0 & 0 & 0 & 0 & 0 & 0 & w13 & w12 \end{pmatrix}_{8\times 8}$$

So the implementation of the first row of mask on the block is derived by defining (6).

$$p = a \times block \times b \quad (6)$$

The DCT representation of "p", "a", "block" and "b" are defined as "P", "A", "BLOCK" and "B", respectively. Lowercase letters represent the matrices in the spatial domain and the capital letters represent the matrices in DCT domain. Equation (6) can be rewritten with (1) and (3) as (7):

$$C. P. C^t = p = a. block. b =$$
$$C^t. A. C. C^t. BLOCK. C. C^t. B. C = C^t. A. BLOCK. B. C \quad (7)$$

According to (7), we can multiply the first row of mask on the block directly in DCT domain by (8).

$$P = A \times BLOCK \times B \quad (8)$$

For implementing the second row of mask on the block, w21, w22 and w23 should slide on the block; therefore matrix "c" is developed as:

$$c = \begin{pmatrix} w22 & w21 & 0 & 0 & 0 & 0 & 0 & 0 \\ w23 & w22 & w21 & 0 & 0 & 0 & 0 & 0 \\ 0 & w23 & w22 & w21 & 0 & 0 & 0 & 0 \\ 0 & 0 & w23 & w22 & w21 & 0 & 0 & 0 \\ 0 & 0 & 0 & w23 & w22 & w21 & 0 & 0 \\ 0 & 0 & 0 & 0 & w23 & w22 & w21 & 0 \\ 0 & 0 & 0 & 0 & 0 & w23 & w22 & w21 \\ 0 & 0 & 0 & 0 & 0 & 0 & w23 & w22 \end{pmatrix}_{8\times 8}$$

The multiplication of the second row of mask on the block is computed directly in DCT domain by (9).

$$Q = BLOCK \times C \quad (9)$$

In the third part, for implementing the third row of mask on the block, w31, w32 and w33 should slide on the block with developing matrix "d". Since the third row of the mask is not related on the last row of the block, the last row of block×d should be zero using multiplication matrix "e" on block. Therefore implementing the third row of mask on the block is computed directly in DCT domain by (10).

$$d = \begin{pmatrix} 0 & 1 & 0 & 0 & 0 & 0 & 0 & 0 \\ 0 & 0 & 1 & 0 & 0 & 0 & 0 & 0 \\ 0 & 0 & 0 & 1 & 0 & 0 & 0 & 0 \\ 0 & 0 & 0 & 0 & 1 & 0 & 0 & 0 \\ 0 & 0 & 0 & 0 & 0 & 1 & 0 & 0 \\ 0 & 0 & 0 & 0 & 0 & 0 & 1 & 0 \\ 0 & 0 & 0 & 0 & 0 & 0 & 0 & 1 \\ 0 & 0 & 0 & 0 & 0 & 0 & 0 & 0 \end{pmatrix}_{8\times 8}$$

$$e = \begin{pmatrix} w32 & w31 & 0 & 0 & 0 & 0 & 0 & 0 \\ w33 & w32 & w31 & 0 & 0 & 0 & 0 & 0 \\ 0 & w33 & w32 & w31 & 0 & 0 & 0 & 0 \\ 0 & 0 & w33 & w32 & w31 & 0 & 0 & 0 \\ 0 & 0 & 0 & w33 & w32 & w31 & 0 & 0 \\ 0 & 0 & 0 & 0 & w33 & w32 & w31 & 0 \\ 0 & 0 & 0 & 0 & 0 & w33 & w32 & w31 \\ 0 & 0 & 0 & 0 & 0 & 0 & w33 & w32 \end{pmatrix}_{8\times 8}$$

$$R = D \times BLOCK \times E \quad (10)$$

Finally the filtering process of the mask in (4) on the 8×8 block can be achieved with the sum of (8), (9) and (10). So the filtering process in the spatial domain (5) is computed directly in DCT domain as (11) which is exactly equal to the spatial domain filtering.

$$F = P + Q + R \quad (11)$$

The achieved mathematical calculations of block filtering in DCT domain will be simpler when the selected filtering mask is symmetric. Because the first row of mask is similar to the third row of mask; similar situation exists for the first column and the third column of mask. So matrix "b" is equal to matrix "e" (so, b=e and B=E) and (11) can be rewritten as (12):

F=P+Q+R=(A×BLOCK×B)+(BLOCK×C)
+(D×BLOCK×E)=(A+D)×BLOCK×E+BLOCK×C=
 AD×BLOCK×E+BLOCK×C         (12)

where AD=A+D.

### C. Creating Replicate Condition

Since the mask has not coverage in the lateral pixels of block, the filtered block does not show the desirable result and zero padding artifacts exist around the edge of the block. To avoid this complication in the spatial domain, as shown in Fig. 1, the 8×8 block is spread to 10×10 block. This operation is done by replicating the value from the nearest lateral pixels. In order to resemble this operation in DCT domain, several matrices are designed to create the border replication condition on edges of block.

The proposed method divides border replication process into several sections including replication the two bottom corners, replication the two upper corners, replication the right, left, up and bottom side of block.

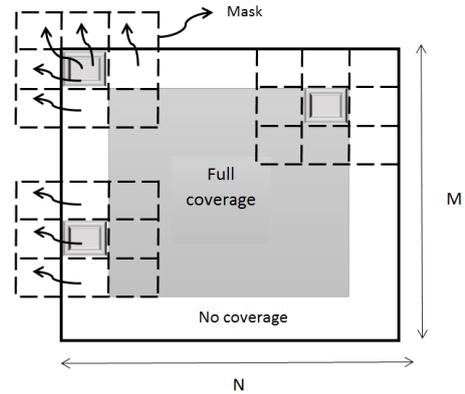

Fig. 1. The schematic of replication condition in a block.

*1) Replication Of Corners:*

Matrix of "bott_cor" (bottom corners) is developed for replicating two bottom corners that the block is multiplied on it. Since only the last row of block×b_c is needed, matrix "b_r" is multiplied on it as bottom or right separator matrix:

$$bott\_cor = \begin{pmatrix} w4+w7+w8 & 0 & 0 & 0 & 0 & 0 & 0 & 0 \\ 0 & 0 & 0 & 0 & 0 & 0 & 0 & 0 \\ 0 & 0 & 0 & 0 & 0 & 0 & 0 & 0 \\ 0 & 0 & 0 & 0 & 0 & 0 & 0 & 0 \\ 0 & 0 & 0 & 0 & 0 & 0 & 0 & 0 \\ 0 & 0 & 0 & 0 & 0 & 0 & 0 & 0 \\ 0 & 0 & 0 & 0 & 0 & 0 & 0 & 0 \\ 0 & 0 & 0 & 0 & 0 & 0 & 0 & w6+w8+I9 \end{pmatrix}_{8\times 8}$$

$$b\_r = \begin{pmatrix} 0 & 0 & 0 & 0 & 0 & 0 & 0 & 0 \\ 0 & 0 & 0 & 0 & 0 & 0 & 0 & 0 \\ 0 & 0 & 0 & 0 & 0 & 0 & 0 & 0 \\ 0 & 0 & 0 & 0 & 0 & 0 & 0 & 0 \\ 0 & 0 & 0 & 0 & 0 & 0 & 0 & 0 \\ 0 & 0 & 0 & 0 & 0 & 0 & 0 & 0 \\ 0 & 0 & 0 & 0 & 0 & 0 & 0 & 0 \\ 0 & 0 & 0 & 0 & 0 & 0 & 0 & 1 \end{pmatrix}_{8\times 8}$$

So the bottom corners replication matrix is calculated directly in DCT domain by (13):

REPL_BOTT_COR= B_R× BLOCK × BOTT_COR   (13)

By bottom corners replication, matrix of "up_cor" (upper corners) and matrix of "u_l" (up or left separator matrix) are developed:

$$\text{up\_cor} = \begin{pmatrix} w1+w2+w4 & 0 & 0 & 0 & 0 & 0 & 0 & 0 \\ 0 & 0 & 0 & 0 & 0 & 0 & 0 & 0 \\ 0 & 0 & 0 & 0 & 0 & 0 & 0 & 0 \\ 0 & 0 & 0 & 0 & 0 & 0 & 0 & 0 \\ 0 & 0 & 0 & 0 & 0 & 0 & 0 & 0 \\ 0 & 0 & 0 & 0 & 0 & 0 & 0 & 0 \\ 0 & 0 & 0 & 0 & 0 & 0 & 0 & 0 \\ 0 & 0 & 0 & 0 & 0 & 0 & 0 & w2+w3+w6 \end{pmatrix}_{8\times 8}$$

$$\text{u\_l} = \begin{pmatrix} 1 & 0 & 0 & 0 & 0 & 0 & 0 & 0 \\ 0 & 0 & 0 & 0 & 0 & 0 & 0 & 0 \\ 0 & 0 & 0 & 0 & 0 & 0 & 0 & 0 \\ 0 & 0 & 0 & 0 & 0 & 0 & 0 & 0 \\ 0 & 0 & 0 & 0 & 0 & 0 & 0 & 0 \\ 0 & 0 & 0 & 0 & 0 & 0 & 0 & 0 \\ 0 & 0 & 0 & 0 & 0 & 0 & 0 & 0 \\ 0 & 0 & 0 & 0 & 0 & 0 & 0 & 0 \end{pmatrix}_{8\times 8}$$

$$\text{up} = \begin{pmatrix} 0 & w1 & 0 & 0 & 0 & 0 & 0 & 0 \\ w3 & w2 & w1 & 0 & 0 & 0 & 0 & 0 \\ 0 & w3 & w2 & w1 & 0 & 0 & 0 & 0 \\ 0 & 0 & w3 & w2 & w1 & 0 & 0 & 0 \\ 0 & 0 & 0 & w3 & w2 & w1 & 0 & 0 \\ 0 & 0 & 0 & 0 & w3 & w2 & w1 & 0 \\ 0 & 0 & 0 & 0 & 0 & w3 & w2 & w1 \\ 0 & 0 & 0 & 0 & 0 & 0 & w3 & 0 \end{pmatrix}_{8\times 8}$$

$$\text{bottom} = \begin{pmatrix} 0 & w7 & 0 & 0 & 0 & 0 & 0 & 0 \\ w9 & w8 & w7 & 0 & 0 & 0 & 0 & 0 \\ 0 & w9 & w8 & w7 & 0 & 0 & 0 & 0 \\ 0 & 0 & w9 & w8 & w7 & 0 & 0 & 0 \\ 0 & 0 & 0 & w9 & w8 & w7 & 0 & 0 \\ 0 & 0 & 0 & 0 & w9 & w8 & w7 & 0 \\ 0 & 0 & 0 & 0 & 0 & w9 & w8 & w7 \\ 0 & 0 & 0 & 0 & 0 & 0 & w9 & 0 \end{pmatrix}_{8\times 8}$$

So the upper corners replication matrix is computed directly in DCT domain by (14).

$$\text{REPL\_UP\_COR} = \text{U\_L} \times \text{BLOCK} \times \text{UP\_COR} \quad (14)$$

*2) Replication The Lateral Side Of Block*

In order to replicate the right side of block, matrix of "right" is developed and is multiplied on block. Then the result is multiplied on "b_r" (bottom or right separator matrix). Similarly, the left side of the block replication is computed using the developed "left" matrix and "up_left" separator matrix. So the right and left sides of block replication are computed directly in DCT domain by (15) and (16), respectively.

$$\text{right} = \begin{pmatrix} 0 & w9 & 0 & 0 & 0 & 0 & 0 & 0 \\ w3 & w6 & w9 & 0 & 0 & 0 & 0 & 0 \\ 0 & w3 & w6 & w9 & 0 & 0 & 0 & 0 \\ 0 & 0 & w3 & w6 & w9 & 0 & 0 & 0 \\ 0 & 0 & 0 & w3 & w6 & w9 & 0 & 0 \\ 0 & 0 & 0 & 0 & w3 & w6 & w9 & 0 \\ 0 & 0 & 0 & 0 & 0 & w3 & w6 & w9 \\ 0 & 0 & 0 & 0 & 0 & 0 & w3 & 0 \end{pmatrix}_{8\times 8}$$

$$\text{left} = \begin{pmatrix} 0 & w7 & 0 & 0 & 0 & 0 & 0 & 0 \\ w1 & w4 & w7 & 0 & 0 & 0 & 0 & 0 \\ 0 & w1 & w4 & w7 & 0 & 0 & 0 & 0 \\ 0 & 0 & w1 & w4 & w7 & 0 & 0 & 0 \\ 0 & 0 & 0 & w1 & w4 & w7 & 0 & 0 \\ 0 & 0 & 0 & 0 & w1 & w4 & w7 & 0 \\ 0 & 0 & 0 & 0 & 0 & w1 & w4 & w7 \\ 0 & 0 & 0 & 0 & 0 & 0 & w1 & 0 \end{pmatrix}_{8\times 8}$$

$$\text{REPL\_RIGHT} = \text{RIGHT} \times \text{BLOCK} \times \text{B\_R} \quad (15)$$

$$\text{REPL\_LEFT} = \text{LEFT} \times \text{BLOCK} \times \text{U\_L} \quad (16)$$

In the last part, we compute the replication matrices for the bottom and upper sides of block. So the upper and bottom sides of block replication are computed directly in DCT domain by (17) and (18), respectively:

$$\text{REPL\_UP} = \text{U\_L} \times \text{BLOCK} \times \text{UP} \quad (17)$$

$$\text{REPL\_BOTT} = \text{B\_R} \times \text{BLOCK} \times \text{BOTTOM} \quad (18)$$

The comprehensive matrix of the border replication condition is achieved as (19):

$$\begin{aligned}\text{REPLICATION\_CONDITION} &= \text{REPL\_BOTT\_COR} + \text{REPL\_UP\_COR} \\ &+ \text{REPL\_RIGHT} + \text{REPL\_LEFT} \\ &+ \text{REPL\_UP} + \text{REPL\_BOTT} \end{aligned} \quad (19)$$

Finally, the filtered block with the border replication is obtained by summing matrix of the filtered block (11) with matrix of border replication condition (19) directly in DCT domain as (20):

$$\text{F\_SYMMETRICAL} = \text{F} + \text{REPLICATION\_CONDITION} \quad (20)$$

When the selected filtering mask is symmetric, we will have UP_COR = BOTT_COR, LEFT = RIGHT and UP=BOTTOM. Therefore, (13) with (14) are integrated together and so on (15) with (16) and (17) with (18). Consequently, the amount of calculations of the border replication condition for the symmetric filtering mask is half of the asymmetric filtering mask in the proposed method.

## III. EXPERIMENT AND RESULT

To assess the validity of the proposed method, the algorithm is performed on 8 ×8 test block in order to filter the block directly in DCT domain. This block and its DCT representation are shown in Fig.2 (a) and Fig.2 (c), respectively. The desired filter is a 3×3 Gaussian mask with σ=1. The mask of the selected filter is shown in Fig.2 (b). In the first part of experiments, filtering process is performed on the block in the spatial domain as ideal result which we desire to reach it. Filtering is performed in the spatial domain with zero padding and then performed with the border replication in block; their results are shown in Fig.2 (d) and Fig.2 (e), respectively. The proposed method prepares the preliminary matrices in DCT domain according to the selected mask of filter. In the first part of the proposed method, block of "F" (14) is computed from DCT representation of test block (Fig.2 (c)). The resulting matrix is equivalent to linear filtering with zero padding in the spatial domain. In the second part of the proposed method, the border replication condition of block in DCT domain is computed and it's added to the matrix obtained in the previous step ("F"). The error between the spatial domain filtering results and the proposed DCT filtering method is negligible. This small difference is due to computation approximation and lossy process for converting image from the spatial domain into DCT domain and vice versa. When the filtered matrix converts to uint8 format, the error reached to zero and the results of the proposed method in DCT domain are exactly the same as the results of the spatial frequency. The image of test block and results are depicted in Fig.2 (j-q). In other experiment we demonstrate the validity of our claims that the proposed method would work for any asymmetric, inseparable and two dimensional filters. Thus, we filtered the test block depicted in Fig.2(a) with mask of the magic square of order 3 depicted in Fig.3(a). Filtering is performed in the spatial domain with boarder replication and then DCT filtering by the proposed method is performed with considering the border replication in the block; their results are shown in Fig.3(b) and Fig.3(c), respectively. The inverse discrete cosine transform of Fig.3(c) is depicted in Fig.3(d).The resulting matrix is equivalent to linear filtering with boarder replication in the spatial domain.

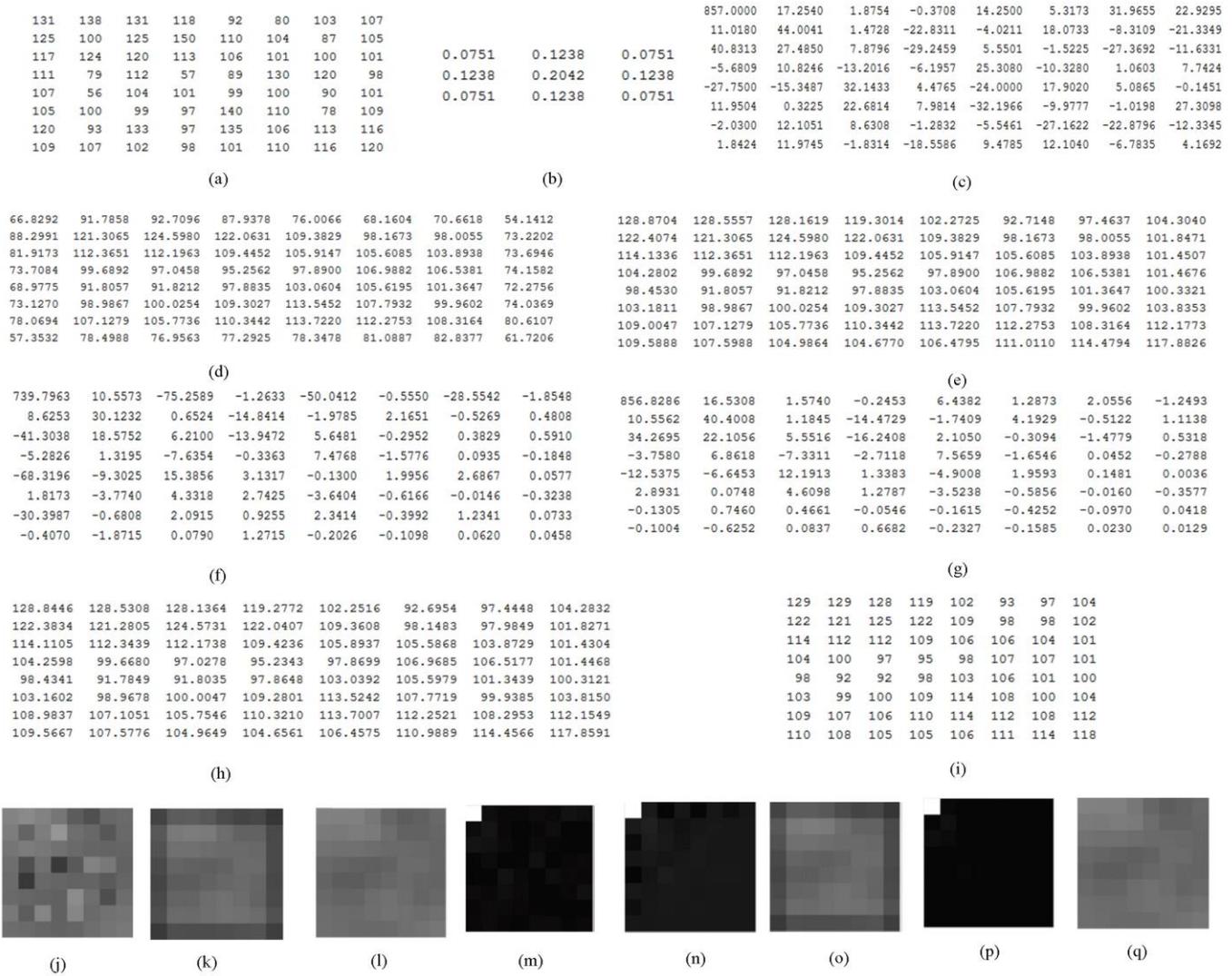

Fig. 2. The first experiment of block filtering in the spatial domain and in DCT domain with the proposed method. (a) Test block matrix. (b) Mask of the selected filter. (c) DCT representation of the test block matrix. (d) Result of the block filtering in the spatial domain with zero padding. (e) Result of the block filtering in the spatial domain with the border replication. (f) Result of the block filtering directly in DCT domain using the proposed method without considering the border replication condition. (g) Result of the block filtering directly in DCT domain using the proposed method with considering the border replication condition. (h) Inverse discrete cosine transform of (g). (i) The uint8 of matrices (e) and (h). (j), (k), (l), (m), (n), (o), (p) and (q) are the displayed image of the matrices (a), (d), (e), (c), (f), idct of (f), (g) and (h), respectively.

## IV. CONCLUSION

We introduced an efficient method for flittering the block directly in DCT domain that has no limitation for selecting the mask of filter such as asymmetric or inseparable or two-dimension. The mathematical computation of the proposed method includes only the same size of block matrices multiplying. These matrices are prepared in DCT domain from the elements of the mask and can be simply modified for any change in the mask. The output results of the proposed filtering method in DCT are exactly same as the spatial domain block filtering by zero padding or border replication in Uint8 format.

REFERENCES

[1] G. K. Wallace, "The JPEG still picture compression standard." *Communications of the ACM*, vol. 34, no. 4, pp. 30-44, 1991.

[2] D. Le Gall, "MPEG: a video compression standard for multimedia applications*," Communications of the ACM*, vol. 34, no. 4, pp. 46-58, 1991.

[3] M. Liou, "Overview of the p×64 kbit/s video coding standard," *Communications of the ACM*, vol. 34, no. 4, pp. 59-63, 1991.

[4] M. A. Naji and A. Aghagolzadeh, "Multi-focus image fusion in DCT domain based on correlation coefficient," in *2015 2nd International Conference on Knowledge-Based Engineering and Innovation (KBEI)*, 2015, pp. 632-639.

[5] M. A. Naji and A. Aghagolzadeh, "A new multi-focus image fusion technique based on variance in DCT domain," in *2015 2nd International Conference on Knowledge-Based Engineering and Innovation (KBEI)*, 2015, pp. 478-484.

[6] W. H. Chen and S. C. Fralick, "Image enhancement using cosine transform filtering*," In Image Sci. Math. Symp.*, Monterey, CA, 1976.

[7] N. N. King and R. J. Clarke, "Lowpass filtering in the cosine transform domain," *International Conference on Communications (ICC '80)*, vol. 2, p. 31.7.1–31.7.5. Seattle, WA, 1980.

[8] B. Chitprasert and K. Rao, "Discrete cosine transform filtering," *1990 International Conference on Acoustics, Speech, and Signal Processing*, ICASSP-90, vol. 3, pp. 1281-1284, 1990.

[9] S. Martucci, "Symmetric convolution and the discrete sine and cosine transforms," *IEEE Transactions on Signal Processing*, vol. 42, no. 5, pp. 1038-1051, 1994.

[10] R. Kresch and N. Merhav, "Fast DCT domain filtering using the DCT and the DST," *IEEE Transactions on Image Processing*, vol. 8, no. 6, pp. 821-833, 1999.

[11] C. Yim, "An Efficient Method for DCT-Domain Separable Symmetric 2-D Linear Filtering" *IEEE Transactions Circuits Systems for Video Technology*, vol. 14, no. 4, pp. 517-521, 2004.

[12] V. G. Reju, S. N. Koh and I. Y. Soon, "Convolution Using Discrete Sine and Cosine Transforms," *IEEE Signal Processing Letters*, vol. 14, no. 7, pp. 445-448, 2007.

[13] K. Suresh and T. V. Sreenivas, "Linear filtering in DCT IV/DST IV and MDCT/MDST domain," *Signal Processing*, vol. 89, no. 6, pp. 1081-1089, 2009.

[14] K. Viswanath, J. Mukherjee and P. K. Biswas, "Image filtering in the block DCT domain using symmetric convolution," *Journal of Visual Communication and Image Representation,* vol. 22, no. 2, pp. 141-152, 2011.

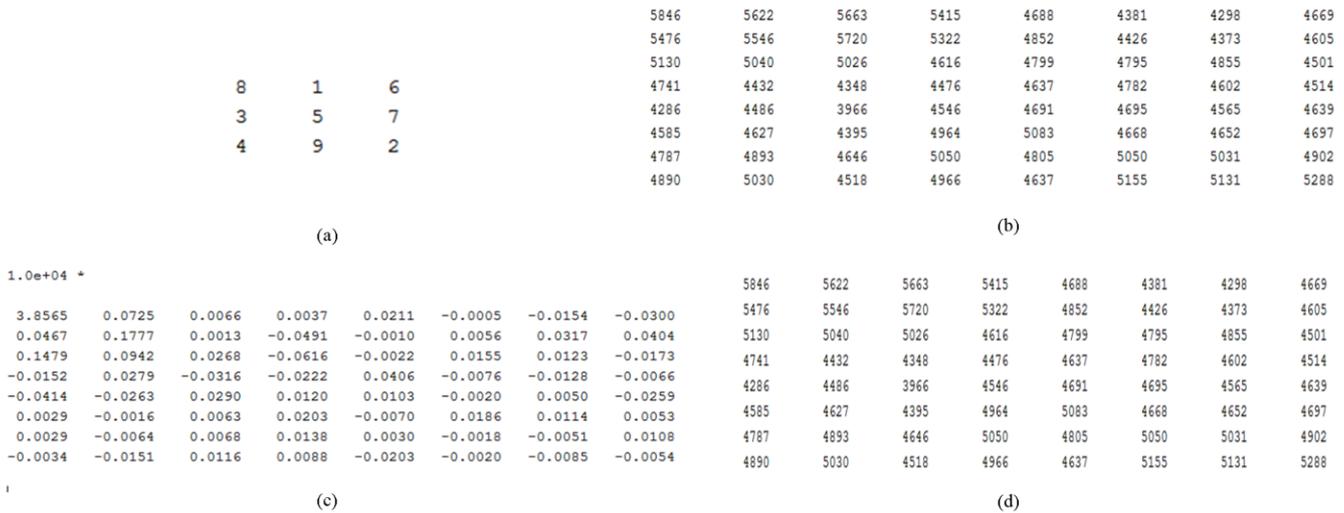

Fig. 3. The second experiment of block filtering in the spatial domain and in DCT domain with the proposed method. (a) The magic square of order 3. (b) Result of the block filtering in the spatial domain with the border replication. (c) Result of the block filtering directly in DCT domain using the proposed method with considering the border replication condition. (d) Inverse discrete cosine transform of (c).